\documentclass{bmvc2k}

\title{Impact of Surface Reflections in \\Maritime Obstacle Detection}

\addauthor{Samed YALÇIN}{yalcinsa21@itu.edu.tr}{1,2}
\addauthor{Hazım Kemal EKENEL}{ekenel@itu.edu.tr/he2244@nyu.edu}{1,3}

\addinstitution{
 Department of Computer Engineering\\
 Istanbul Technical University\\
 Istanbul, Turkey
}
\addinstitution{
 Aselsan A.Ş.\\
 Ankara, Turkey
}
\addinstitution{
 Division of Engineering\\
 NYU Abu Dhabi\\
 Abu Dhabi, UAE
}

\runninghead{Yalçın, Ekenel}{Impact of Surface Reflections}

\usepackage{colortbl}
\usepackage{algorithm}
\usepackage{algpseudocode}

\begin{document}

\maketitle

\begin{abstract}
Maritime obstacle detection aims to detect possible obstacles for autonomous driving of unmanned surface vehicles.
In the context of maritime obstacle detection, the water surface can act like a mirror on certain circumstances, causing reflections on imagery.
Previous works have indicated surface reflections as a source of false positives for object detectors in maritime obstacle detection tasks.
In this work, we show that surface reflections indeed adversely affect detector performance.
We measure the effect of reflections by testing on two custom datasets, which we make publicly available. The first one contains imagery with reflections, while in the second reflections are inpainted. We show that the reflections reduce mAP by 1.2 to 9.6 points across various detectors.
To remove false positives on reflections, we propose a novel filtering approach named Heatmap Based Sliding Filter. We show that the proposed method reduces the total number of false positives by 34.64\% while minimally affecting true positives. We also conduct qualitative analysis and show that the proposed method indeed removes false positives on the reflections. The datasets can be found on \href{https://github.com/SamedYalcin/MRAD}{https://github.com/SamedYalcin/MRAD}.
\end{abstract}

\section{Introduction}
\label{sec:intro}
Research on unmanned and autonomous vehicles is accelerating day by day. While these studies have predominantly focused on unmanned aerial vehicles, recent years have seen an increasing interest in unmanned surface vehicles (USV) as well ~\cite{macvi1}~\cite{macvi2}~\cite{MODS}~\cite{ABOShips}. One of the critical capabilities of these unmanned vehicles is autonomous driving. A key factor influencing this capability is the obstacle detection.

Obstacles are objects on the water surfaces that USVs can collide with, making their accurate detection crucial for the performance of autonomous navigation. Various computer vision methods, such as Convolutional Neural Networks (CNNs)~\cite{alexnet}, have been employed to tackle this challenge. More recently, transformer-based detectors ~\cite{ViT}~\cite{codetr} have emerged, exceeding the capabilities of CNN-based approaches.

However, the complexity of detecting obstacles on the water surfaces is compounded by the reflective nature of water, which can act as a mirror under certain weather conditions. This leads to the appearance of object reflections in the images captured by USV cameras, as shown in Figure \ref{fig:dataset}(a). These reflections can be mistakenly identified as obstacles, increasing the number of false positives and potentially hindering critical tasks such as path planning. 

Previous studies~\cite{macvi1}~\cite{macvi2} have indicated that reflections on the water surfaces can be a source of false positives, but the quantitative effect of these reflections has not been measured in isolation. In this study, we visually confirm such false positives on reflections and quantitatively measure the impact of reflections in isolation, demonstrating that reflections do indeed negatively affect object detectors. Furthermore, we propose a novel filtering approach, named Heatmap Based Sliding Filter (HBSF) to eliminate such false positives. We show that the proposed filter reduces the number of false positives while minimally affecting true positives. We present a qualitative analysis of the proposed filtering approach, showing its ability to remove false positives on reflections. We present the following contributions:
\vspace{-0.4em}
\begin{itemize}
\itemsep-0.4em
\item We collect a dataset of images captured from a perspective similar to that of a USV containing reflections, along with a twin dataset where the reflections have been removed by inpainting allowing analysis of the impact of reflections.
\item We evaluate various CNN models (RCNN~\cite{RCNN} and YOLO~\cite{YOLO} variants) as well as a state-of-the-art transformer based detector, Co-DETR~\cite{codetr}, on the constructed datasets and demonstrate that reflections are indeed a significant negative factor.
\item We propose a novel filter, named HBSF, that removes false positives on surface reflections with minimal effect on the true positives.
\end{itemize}

\section{Related Works}
\label{sec:relatedworks}
The impact of reflections on object detection has been a notable challenge in various autonomous systems, particularly in maritime environments. In 2023, the 1st Workshop on Maritime Computer Vision (MaCVi)~\cite{macvi1} hosted a USV based obstacle detection challenge. The challenge was based on the dataset MODS~\cite{MODS}, a USV oriented object detection and obstacle segmentation benchmark. The challenge identified surface reflections as one of the difficulties of maritime obstacle detection. This year, the 2nd Workshop on Maritime Computer Vision~\cite{macvi2} held the same challenge with a newly created dataset named LaRS~\cite{LaRS} which consists of diverse scenes including various levels of reflection. In the challenge, Kiefer et al.~\cite{macvi2} noted that reflections still posed a significant challenge to both USV-based detection and segmentation tasks. They observed a significant drop in F1 score from 80\% in moderately reflective scenes to below 40\% in highly reflective scenes during the obstacle segmentation challenge, highlighting the necessity for robust detection systems capable of distinguishing between real objects and their reflections.

Prior work has primarily focused on removing reflections from glossy surfaces of solid materials such as window glass~\cite{sirr1}\cite{sirr2}\cite{sirr3}. Wang et al., in their recent study, introduced the Reflected Object Detection Dataset (RODD)~\cite{rodd}, along with a novel approach that trains detectors to classify whether an object is real or reflected. However, these methods are generally designed for reflections on solid surfaces and may not effectively address the challenges posed by reflections on water bodies. Our work aims to bridge this gap by specifically focusing on maritime reflections, an area that has received less attention in the literature. Unlike prior approaches, we concentrate on removing false positives on reflections on water surfaces.

\section{Datasets}
\label{sec:datasets}
The maritime object detection domain has seen a surge of datasets in recent years.
In 2022, Bovcon et al. published MODS~\cite{MODS}, a benchmark for USV-oriented obstacle segmentation and detection. To reflect the difficulty of maritime domain, Bovcon et al. obtained samples using an image acquisition system mounted on a small sized USV, amounting to  8K images and  64K objects. Later in 2023, \v{Z}ust et al. published LaRS (Lakes, Rivers, and Seas)~\cite{LaRS}, another USV-based maritime obstacle detection benchmark, extending the diversity of the scenes compared to previous datasets. Notably, \v{Z}ust et al. labelled each image with 19 different global attributes, e.g. lighting conditions, reflection levels etc., allowing analysing the effect of scene conditions in maritime perception tasks.

\begin{figure}[t]
\begin{tabular}{ccc}
\bmvaHangBox{\fbox{\includegraphics[width=3.25cm]{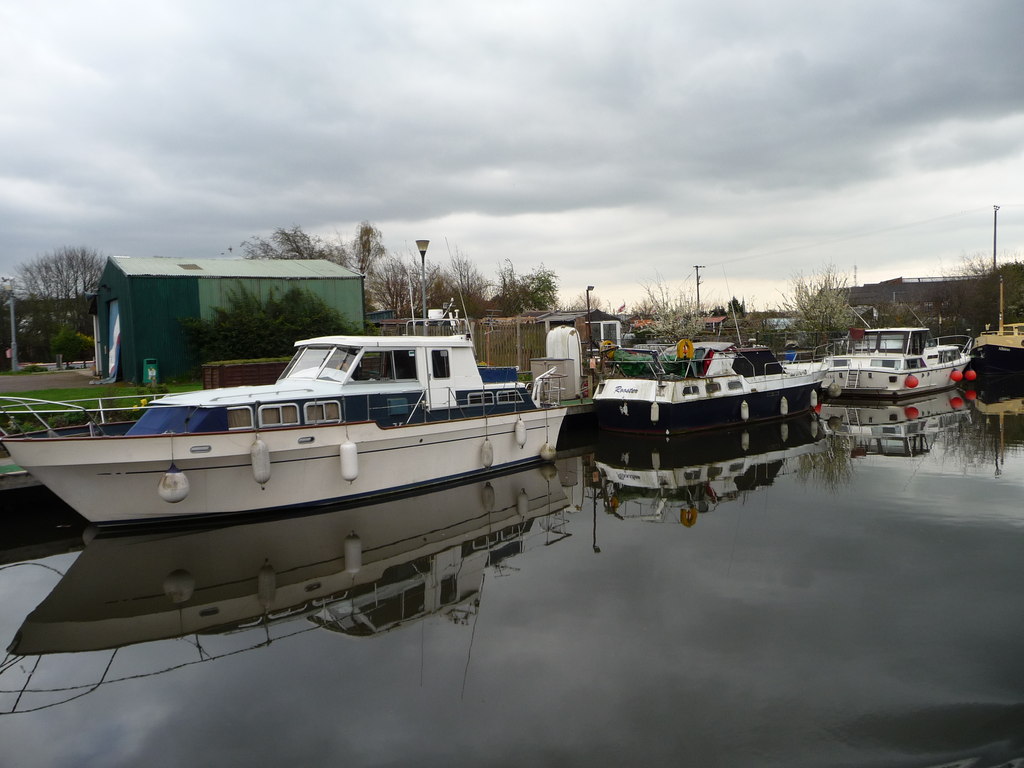}}}&
\bmvaHangBox{\fbox{\includegraphics[width=3.25cm]{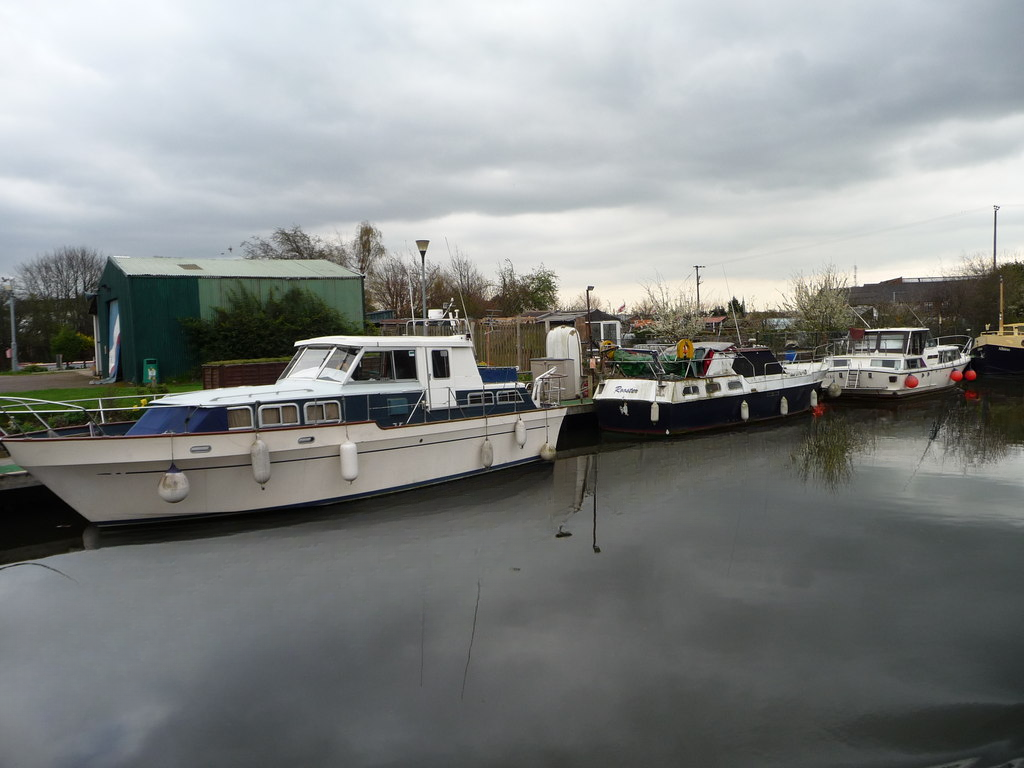}}}&
\bmvaHangBox{\fbox{\includegraphics[width=4.2cm]{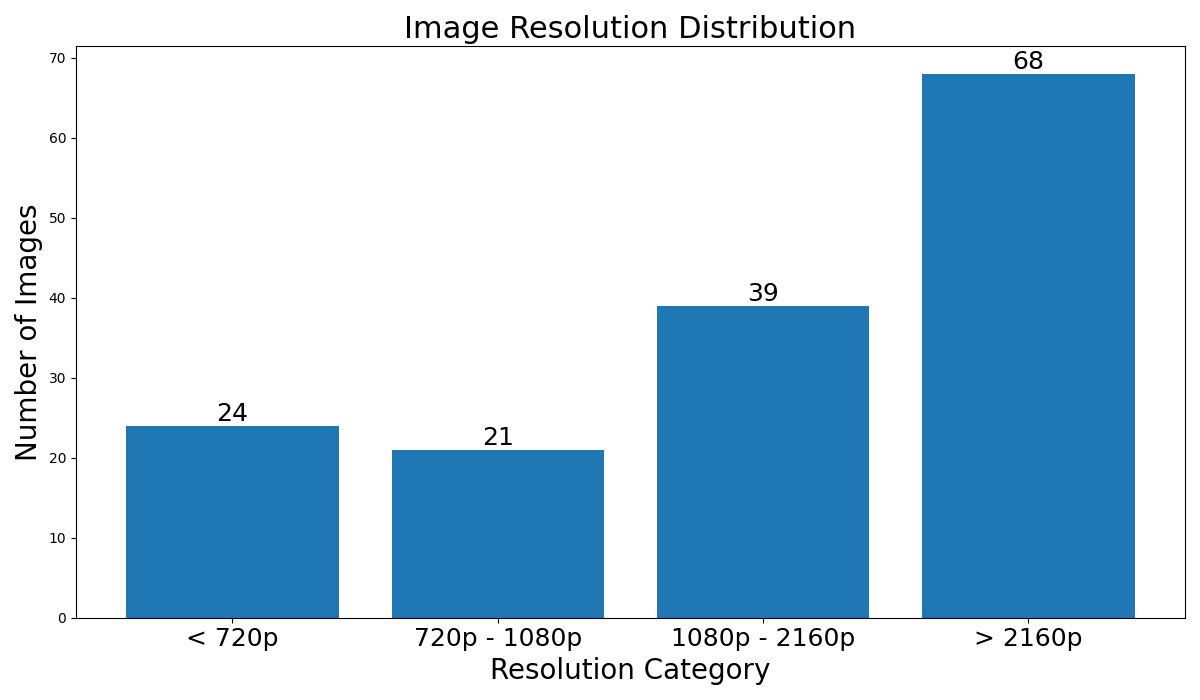}}}\\
(a)&(b)&(c)
\end{tabular}
\caption{
(a) Original sample from the dataset;
(b) reflections of the boats in (a) are inpainted;
(c) dataset image size distribution. Best viewed zoomed in.
[Source:
(a) - Boats moored on the River Don New Cut by Christine Johnstone, licensed under CC BY-SA 2.0;
(b) - Modified from (a) under the same license.]
}
\label{fig:dataset}
\end{figure}

Out of these datasets, MODS was used in the 1st MaCVi Workshop and LaRS was used in the 2nd MaCVi workshop. Based on the results of the USV-based obstacle detection challenges in the MaCVi workshops, we decided to work on the effect of surface reflections.
While MODS and LaRS are USV-based, the number of highly reflective scenes in these datasets are limited. Images of highly reflective scenes not always capture the reflection of the objects in a prominent way, and image quality can be lacking. For this reason, we opted to construct a small dataset consisting of high quality images of highly reflective scenes. We think, for downstream tasks like path planning, boats will be the main consideration due to their sizes which pose a bigger risk for collision compared to smaller objects like buoys. Therefore, we focused on boats. We collected 152 images of boats in various scenes of various sizes, from $640\times480$ to above 8K. Size distribution is shown in Figure \ref{fig:dataset}(c). Images were collected from Wikimedia Commons, using keywords like “boat” and “reflection”. All the images are licensed under Creative Commons licence. We also created a duplicate of this dataset and inpainted the reflections using LaMa~\cite{lama}. A sample from the dataset and the version where the reflections of boats are inpainted are shown in Figure \ref{fig:dataset} (a) and (b). In total, we annotated 490 boats. According to COCO~\cite{MSCOCO} definition of small, medium, and large, there are 36, 111, and 343 objects, respectively.

\section{Impact of Surface Reflections}
\label{sec:analysis}
The reflections are identified as a source of false positives in~\cite{macvi1}~\cite{macvi2}. Since domain adapted detectors demonstrate this behaviour in~\cite{macvi1}~\cite{macvi2}, we start by looking at the problem with a broader perspective. Analysing the effect of reflections on detector performance requires isolating the reflection variable. One option might be to curate two datasets, one with only images with reflections and one without, each sufficiently large and similar in distribution. This option comes with its own challenges: budget and ensuring the distributions of the datasets are similar so that the only difference would be the reflections. The other option, which we follow, is to simply inpaint the reflections.

Since in real life, input images will inevitably include reflections, we use our dataset as a benchmark and measure various detectors' performance. Then, we repeat the same process using the inpainted version of the dataset. Performances of detectors are then compared to analyse the effect of reflections. Following, we utilize transfer learning to see if models adapted to the maritime domain behave differently. Finally, we experiment with different, more complex backbone structures and analyse their effect.

To assess model performance, standard metrics such as MS COCO~\cite{MSCOCO} mAP are utilized. True positive (TP) and false positive (FP) counts are analysed where necessary to obtain a more fine-grained picture. The average IoU of proposals with available ground truths is computed to see if reflections affect the tightness of the proposals. Visual analysis is also conducted. We expand on the experiments and metrics in Section \ref{sub:experiments}. 

We also propose a novel filtering approach to remove proposals on reflections, named HBSF, which we expand on in Section \ref{sec:hbefhbsf}.

\subsection{Experiments}
\label{sub:experiments}
Three different experiments are conducted. In Section \ref{subsec:experiment1}, various pretrained detectors are analysed on our dataset to assess sensitivity of different architectures to reflections. In Section \ref{subsec:experiment2}, transfer learning is applied to adapt models to the maritime domain. The fine-tuned model is analysed on our dataset. In Section \ref{subsec:experiment3}, different backbone structures are used to evaluate how different feature extractors are affected.

All experiments are conducted on the Kaggle platform with single Intel(R) Xeon(R) CPU @ 2.00GHz 2C/4T, 29GB CPU RAM, single NVidia P100 GPU and 16GB VRAM. All models are obtained from OpenMMLab's MMDetection V2 toolbox~\cite{mmdetection}.

\subsubsection{Model Sensitivity}
\label{subsec:experiment1}
We analyse both single stage and two stage detectors, as well as a transformer based approach, and assess the sensitivity of different architectures.
For single stage detectors, YOLO~\cite{YOLO} based YOLOv3~\cite{YOLOV3}, YOLOf~\cite{YOLOf} and YOLOx~\cite{YOLOx} are used.
For two stage detectors RCNN~\cite{RCNN} based FRCNN~\cite{FasterRCNN}, CRCNN~\cite{CascadeRCNN}, HTC~\cite{HTC} and DetectoRS~\cite{DetectoRS} are used.
As transformer based detector, state-of-the-art Co-DETR~\cite{codetr} is used.
All models are pretrained on MS COCO and no additional training is applied.
Since MS COCO includes a boat category, we can do inference without any alterations to pretrained models.
All models are evaluated on our datasets. We then analyse the mAP scores and average IoUs of the proposals.
Table \ref{table:experiment1} shows the mAP scores. We observe the following:

\begin{table}[t]
  \rowcolors{2}{gray!25}{white}
  \centering
  \begin{tabular}{ccccc}
    Reflections                     & Model                     & mAP[0.50:0.95] & mAP[0.50]  & mAP[0.75]      \\
    Original                        & FRCNN                     & 0.231          & 0.503      & 0.189          \\
    Inpainted                       & FRCNN                     & 0.314          & 0.574      & 0.316          \\
    Original                        & CRCNN                     & 0.288          & 0.575      & 0.245          \\
    Inpainted                       & CRCNN                     & 0.384          & 0.652      & 0.391          \\
    Original                        & HTC                       & 0.313          & 0.594      & 0.290          \\
    Inpainted                       & HTC                       & 0.393          & 0.658      & 0.397          \\
    Original                        & DetectoRS                 & 0.450          & 0.715      & 0.464          \\
    Inpainted                       & DetectoRS                 & 0.462          & 0.727      & 0.481          \\
    Original                        & YOLOv3                    & 0.208          & 0.509      & 0.147          \\
    Inpainted                       & YOLOv3                    & 0.268          & 0.573      & 0.226          \\
    Original                        & YOLOf                     & 0.243          & 0.549      & 0.186          \\
    Inpainted                       & YOLOf                     & 0.338          & 0.603      & 0.324          \\
    Original                        & YOLOx                     & 0.366          & 0.645      & 0.360          \\
    Inpainted                       & YOLOx                     & 0.421          & 0.689      & 0.420          \\
    Original                        & CoDETR                    & 0.601          & 0.811      & 0.652          \\
    Inpainted                       & CoDETR                    & 0.583          & 0.811      & 0.618          \\
    Inpainted(ZITS)                 & CoDETR                    & 0.581          & 0.813      & 0.636          
  \end{tabular}
\caption{\textsc{Sensitivity Analysis Results}}
  \label{table:experiment1}
\end{table}

\begin{table}[t]
  \rowcolors{2}{gray!25}{white}
  \centering
  \begin{tabular}{cccccc}
    Reflections       & Model      & IoU[0.05]  & IoU[0.3]  & IoU[0.5] & IoU[0.7]        \\
    Original          & FRCNN      & 0.278      & 0.396      & 0.462     & 0.538          \\
    Inpainted         & FRCNN      & 0.293      & 0.423      & 0.493     & 0.582          \\
    Original          & CRCNN      & 0.284      & 0.483      & 0.568     & 0.634          \\
    Inpainted         & CRCNN      & 0.297      & 0.520      & 0.617     & 0.688          \\
    Original          & HTC        & 0.286      & 0.480      & 0.555     & 0.626          \\
    Inpainted         & HTC        & 0.302      & 0.520      & 0.599     & 0.678          \\
    Original          & DetectoRS  & 0.320      & 0.537      & 0.607     & 0.679          \\
    Inpainted         & DetectoRS  & 0.333      & 0.556      & 0.617     & 0.684          \\
    Original          & YOLOv3     & 0.393      & 0.585      & 0.642     & 0.679          \\
    Inpainted         & YOLOv3     & 0.421      & 0.621      & 0.674     & 0.722          \\
    Original          & YOLOf      & 0.212      & 0.469      & 0.714     & 0.729          \\
    Inpainted         & YOLOf      & 0.214      & 0.504      & 0.761     & 0.791          \\
    Original          & YOLOx      & 0.391      & 0.609      & 0.688     & 0.749          \\
    Inpainted         & YOLOx      & 0.401      & 0.630      & 0.718     & 0.781          \\
    Original          & CoDETR     & 0.225      & 0.593      & 0.761     & 0.877          \\
    Inpainted         & CoDETR     & 0.233      & 0.596      & 0.756     & 0.864          
  \end{tabular}
\caption{\textsc{Average IoUs of proposals at different score thresholds}}
  \label{table:experiment1_iou}
\end{table}

\begin{figure}[h]
\begin{tabular}{cc}
\bmvaHangBox{\fbox{\includegraphics[width=5.5cm]{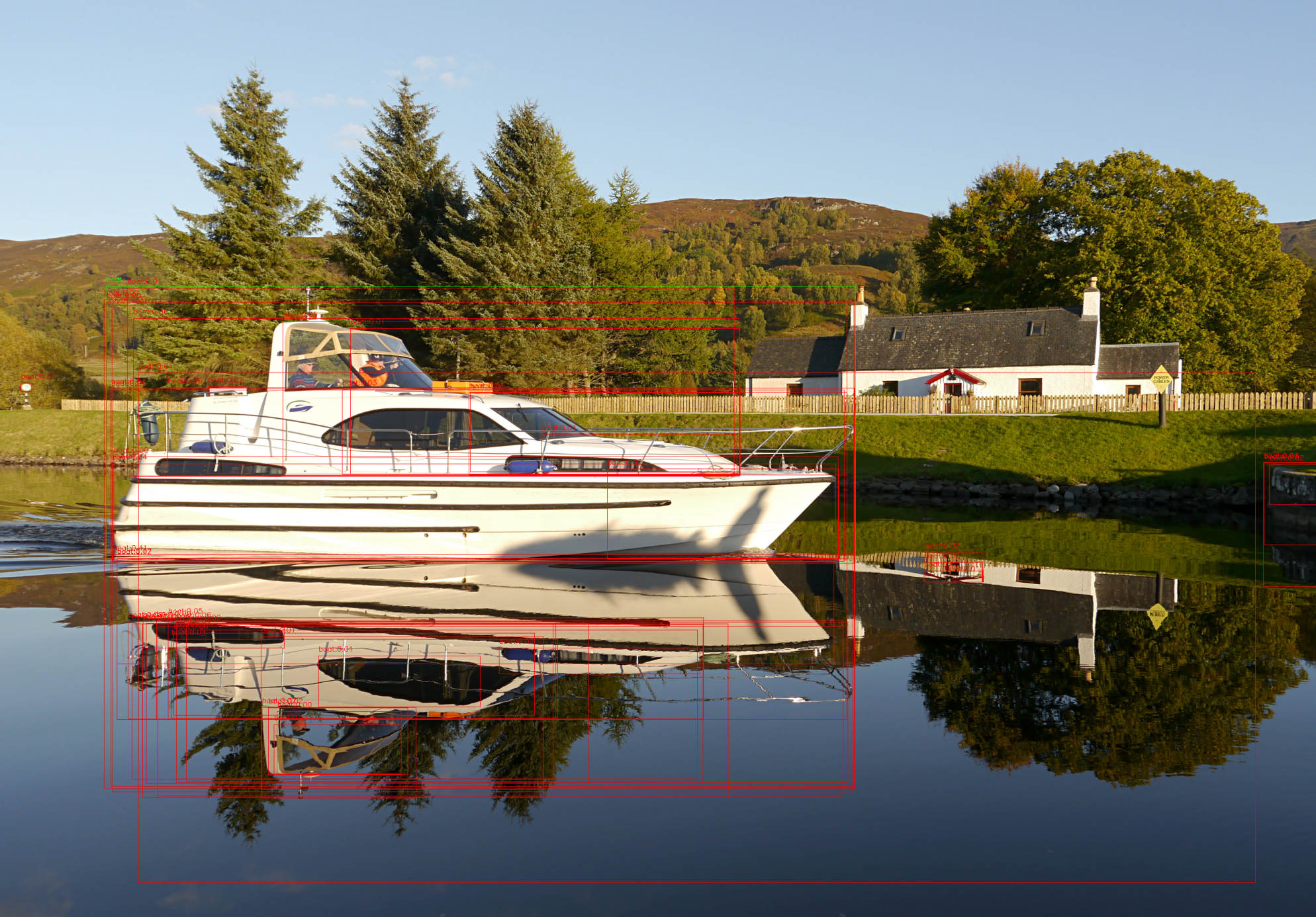}}}&
\bmvaHangBox{\fbox{\includegraphics[width=5.5cm]{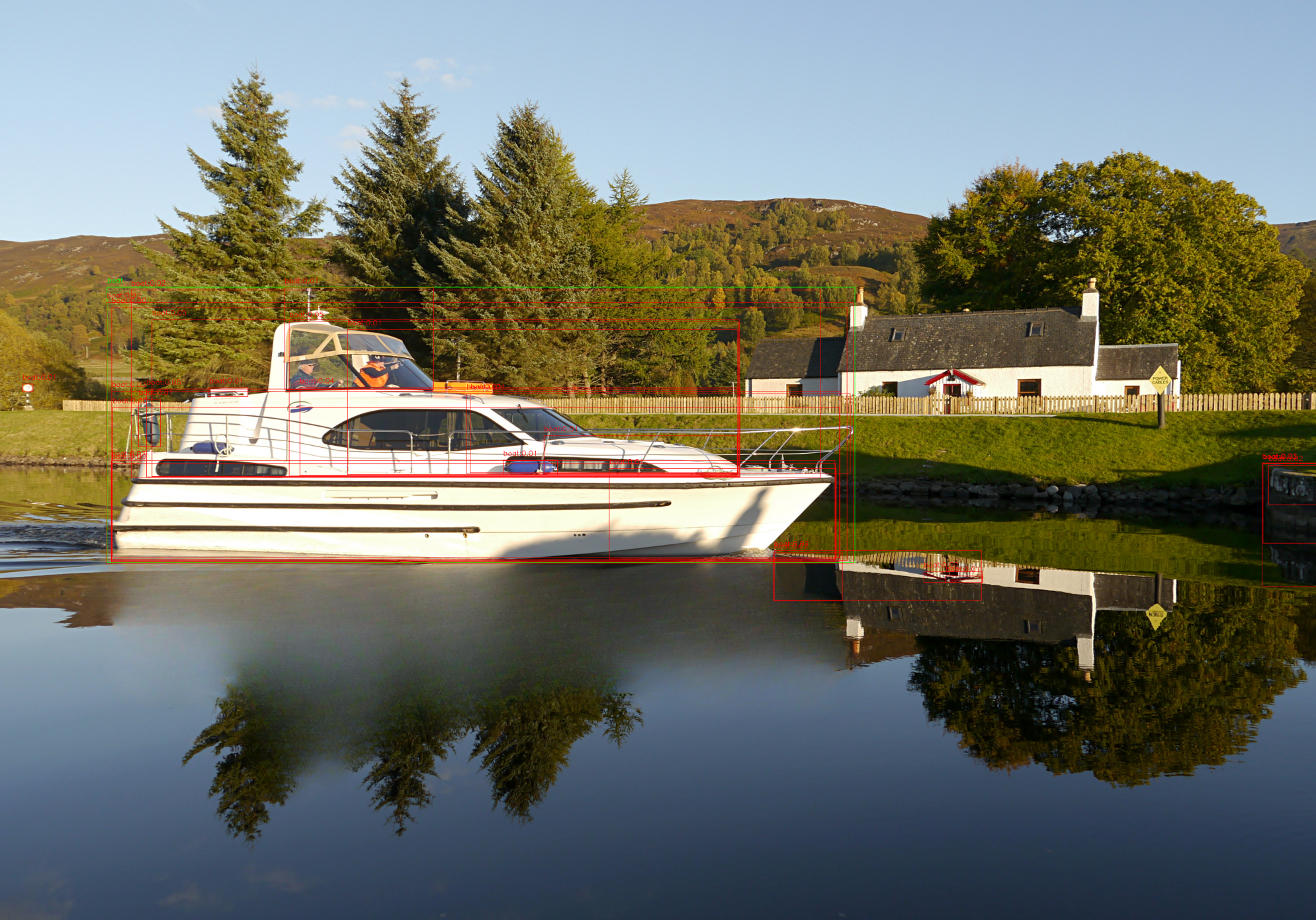}}}\\
(a)&(b)
\end{tabular}
\caption{(a) Sample false positive due to surface reflection; (b) Inpainted sample showcasing removal of false positives on reflections. Both proposals are from Co-DETR.
[Source: Modified from the photo "Boat passing bridge keeper's cottage, Aberchalder by Craig Wallace" under the same license, CC BY-SA 2.0]}
\label{fig:fponreflection}
\end{figure}

First, all CNN-based architectures experience a significant increase in mAP when reflections are removed.
Furthermore, mAP@0.75 increases more than mAP@0.50.
Visual inspection indicates that the presence of reflections introduces false positives on reflections along with proposals that encompass both the reflection and the object itself, counting towards FP count.
Figure \ref{fig:fponreflection} showcases this behaviour.
Combined proposals also reduce TP count at higher thresholds.
When reflections are removed, FP@0.50 (false positive count at IoU threshold of 0.5) reduces from 7571 to 6958 while TP@0.50 increases from 390 to 404 for FRCNN.
Similarly, FP@0.75 reduces from 7792 to 7145 while TP@0.75 increases from 169 to 244, supporting our hypothesis.

\begin{figure}[h]
\centering
\begin{tabular}{cc}
\bmvaHangBox{\fbox{\includegraphics[width=3.4cm]{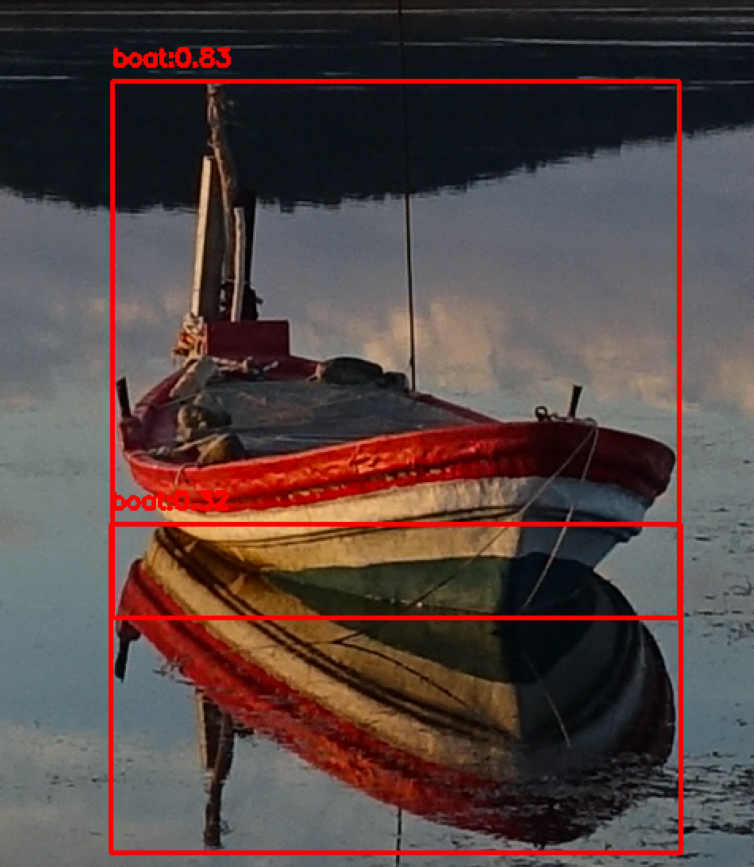}}}&
\bmvaHangBox{\fbox{\includegraphics[width=3.8cm]{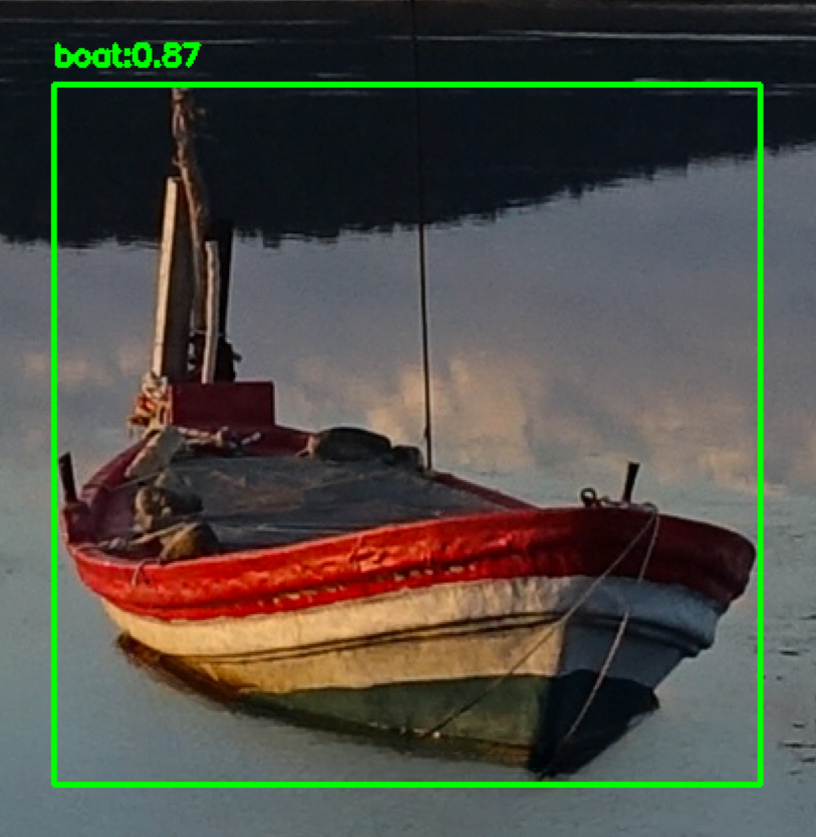}}}\\
(a)&(b)
\end{tabular}
\caption{(a) Original sample; (b) inpainted sample. Notice how inpainting enlarges the object downward. Proposals are from Co-DETR.
[Source: Modified from the photo by Dimitrios Tzortzis under the same license, CC BY-SA 4.0]
}
\label{fig:inpaintexpand}
\end{figure}

Second, Co-DETR \emph{seems} to be negatively affected.
Looking at FP and TP counts, we notice  11.9\% reduction in FP count for all IoU thresholds, indicating there were false positives on the reflections.
Visual analysis confirms this, as shown in Figure \ref{fig:fponreflection}.
However, at higher thresholds, TP count seems to reduce.
TP@0.7 and below is affected minimally, but TP@0.8 reduces by 2.47\% and TP@0.9 reduces by 5.3\%.
We change the inpainting algorithm to ZITS~\cite{zits} to see how Co-DETR behaves with different inpainting approaches.
This time we observe an increase in mAP@0.5 and mAP@0.75 compared to the results obtained on the inpainted images generated by LaMa.
We hypothesize that artefacts left by the inpainting algorithms affect how Co-DETR behaves.
Visual analysis shown in Figure \ref{fig:inpaintexpand} demonstrates that inpainting algorithms may expand the objects downward, potentially reducing IoU of proposals and degrading model performance.

In Table \ref{table:experiment1_iou} we present average IoUs of each models' proposals on our datasets. We apply different score thresholds, noted in brackets, and only consider proposals above this threshold to obtain a more fine-grained view. We notice that CNN-based detectors experience significant increases in IoU, especially for more confident proposals. This behaviour is different in Co-DETR. According to the result of IoU[0.05], the intersection of proposals is increased. However, more confident proposals become looser, which is expected as inpainting expands some objects downward, as illustrated in Figure \ref{fig:inpaintexpand}.

\subsubsection{Domain Adaptation}
\label{subsec:experiment2}
\begin{figure}[h]
\centering
\begin{tabular}{cc}
\bmvaHangBox{\fbox{\includegraphics[width=6cm]{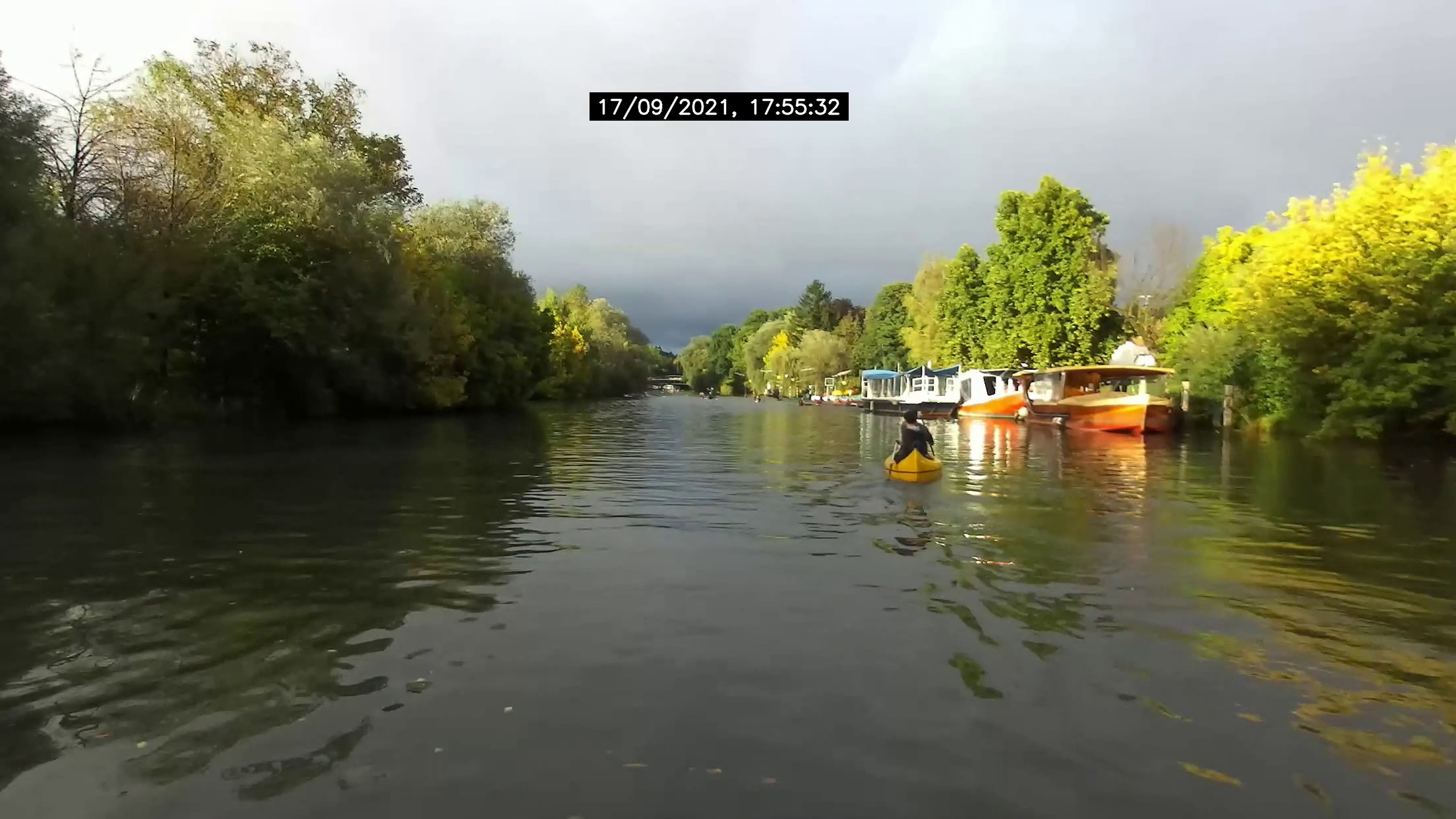}}}&
\bmvaHangBox{\fbox{\includegraphics[width=5.05cm]{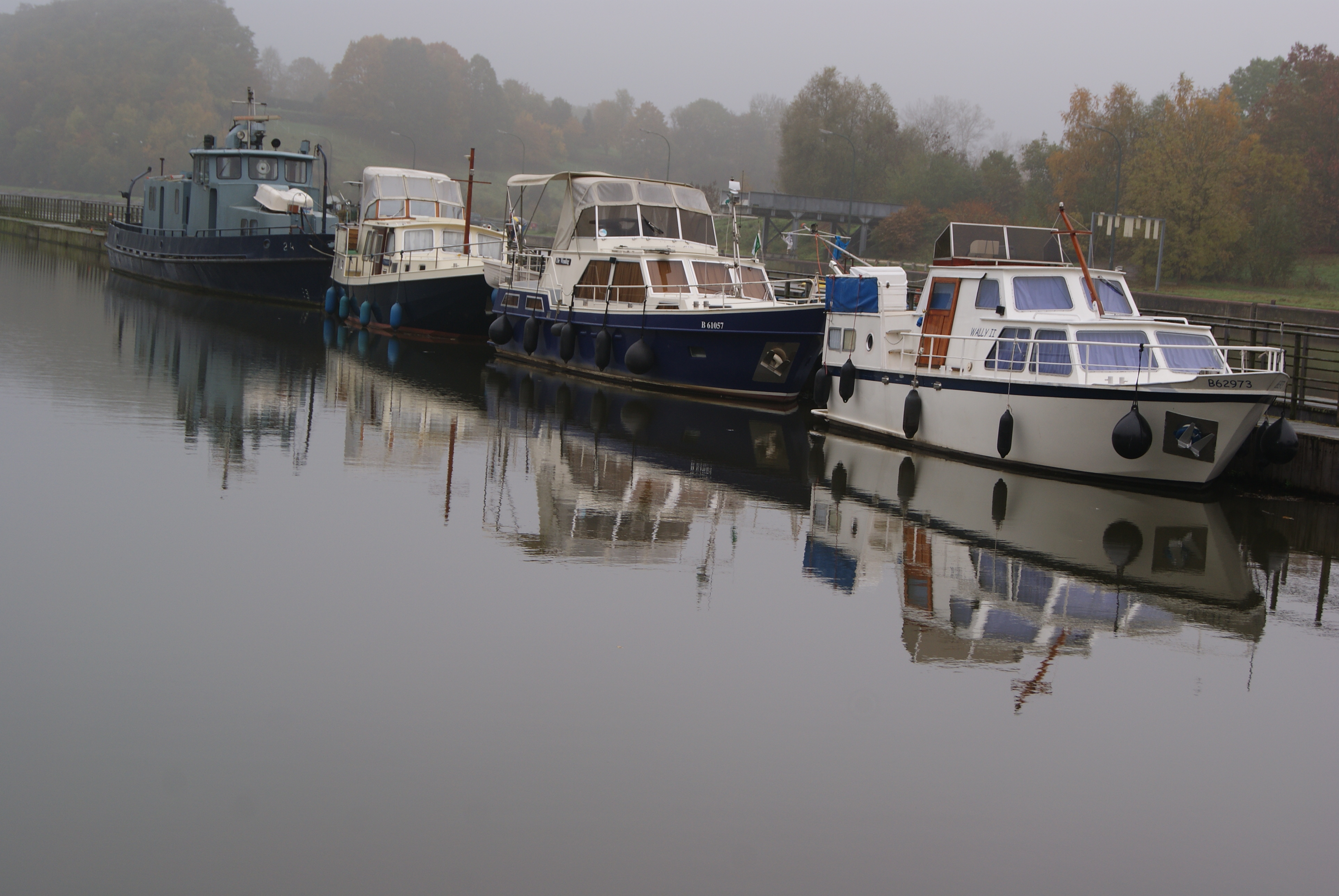}}}\\
(a)&(b)
\end{tabular}
\caption{(a) An example from LaRS dataset. (b) An example from our dataset. Notice the differences is scenery and quality.
[Source: (b) - Photo by Glaurent, licensed under CC BY 4.0]}
\label{fig:lars_v_ours}
\end{figure}

For this experiment, we fine-tune FRCNN using the LaRS dataset and evaluate on our dataset. The results are presented in Table \ref{table:experiment2}. We observe that with transfer learning, removing reflections grants us 4\% mAP improvement. The improvement increases to 12.1\% for mAP@0.75. We notice overall performance reduces compared to the pretrained model, but this is not out of the ordinary considering that model now expects data similar to LaRS. As shown in \ref{fig:lars_v_ours}, although both LaRS and our dataset include similar objects, scenes between datasets are not exactly alike, explaining the degradation. Since we observe a trend similar to results in Section \ref{subsec:experiment1}, we conclude that even with domain adaptation, reflections are a source of false positives. This is also inline with findings in~\cite{macvi1}~\cite{macvi2}.

\begin{table}[h]
  \rowcolors{2}{gray!25}{white}
  \centering
  \begin{tabular}{ccccc}
    Reflections                     & Model                     & mAP[0.50:0.95] & mAP[0.50]  & mAP[0.75]      \\
    Original                        & FRCNN                     & 0.205          & 0.445      & 0.169          \\
    Inpainted                       & FRCNN                     & 0.245          & 0.490      & 0.290          \\
  \end{tabular}
\caption{\textsc{Domain Adaptation Results}}
  \label{table:experiment2}
\end{table}

\subsubsection{Backbone Complexity}
\label{subsec:experiment3}
Finally, we analyse the performance of different backbones. For this purpose, we use FRCNN with 3 different backbones ResNet50~\cite{ResNet}, ResNet101 and ResNeXt101~\cite{ResNeXt} all with the FPN~\cite{FPN} module. Results are presented in Table \ref{table:experiment3}. In line with previous experiments, we observe an increase in mAP scores when reflections are removed for all three backbone architectures. Similarly, mAP@0.75 increases more than mAP@0.5. Unlike other backbones, ResNeXt101 shows only a 0.008 point increase on mAP@0.50. We visually analyse the proposals to understand this behaviour. What we notice is that different from ResNet50 and ResNet101, ResNeXt101 generates fewer proposals that extend downward due to reflections. Therefore, at lower IoU thresholds, the precision-recall curve does not change significantly. However, at higher IoU thresholds, removal of reflections reduces FPs on reflections, increasing the mAP.
 
\begin{table}[h]
  \rowcolors{2}{gray!25}{white}
  \centering
  \begin{tabular}{cccccc}
    Reflections                     & Model                     & Backbone                     & mAP[0.50:0.95] & mAP[0.50]  & mAP[0.75]      \\
    Original                        & FRCNN                     & R-50-FPN                     & 0.260          & 0.567      & 0.211          \\
    Inpainted                       & FRCNN                     & R-50-FPN                     & 0.348          & 0.639      & 0.332          \\
    Original                        & FRCNN                     & R-101-FPN                    & 0.302          & 0.592      & 0.279          \\
    Inpainted                       & FRCNN                     & R-101-FPN                    & 0.375          & 0.661      & 0.377          \\
    Original                        & FRCNN                     & RX-101-FPN-32                & 0.389          & 0.690      & 0.390          \\
    Inpainted                       & FRCNN                     & RX-101-FPN-32                & 0.440          & 0.698      & 0.480          \\
  \end{tabular}
\caption{\textsc{Performance Change on Different Backbones}}
  \label{table:experiment3}
\end{table}

\section{Heatmap Based Sliding Filter}
\label{sec:hbefhbsf}
\begin{figure}[h]
\centering
\begin{tabular}{cc}
\bmvaHangBox{\fbox{\includegraphics[width=4.2cm]{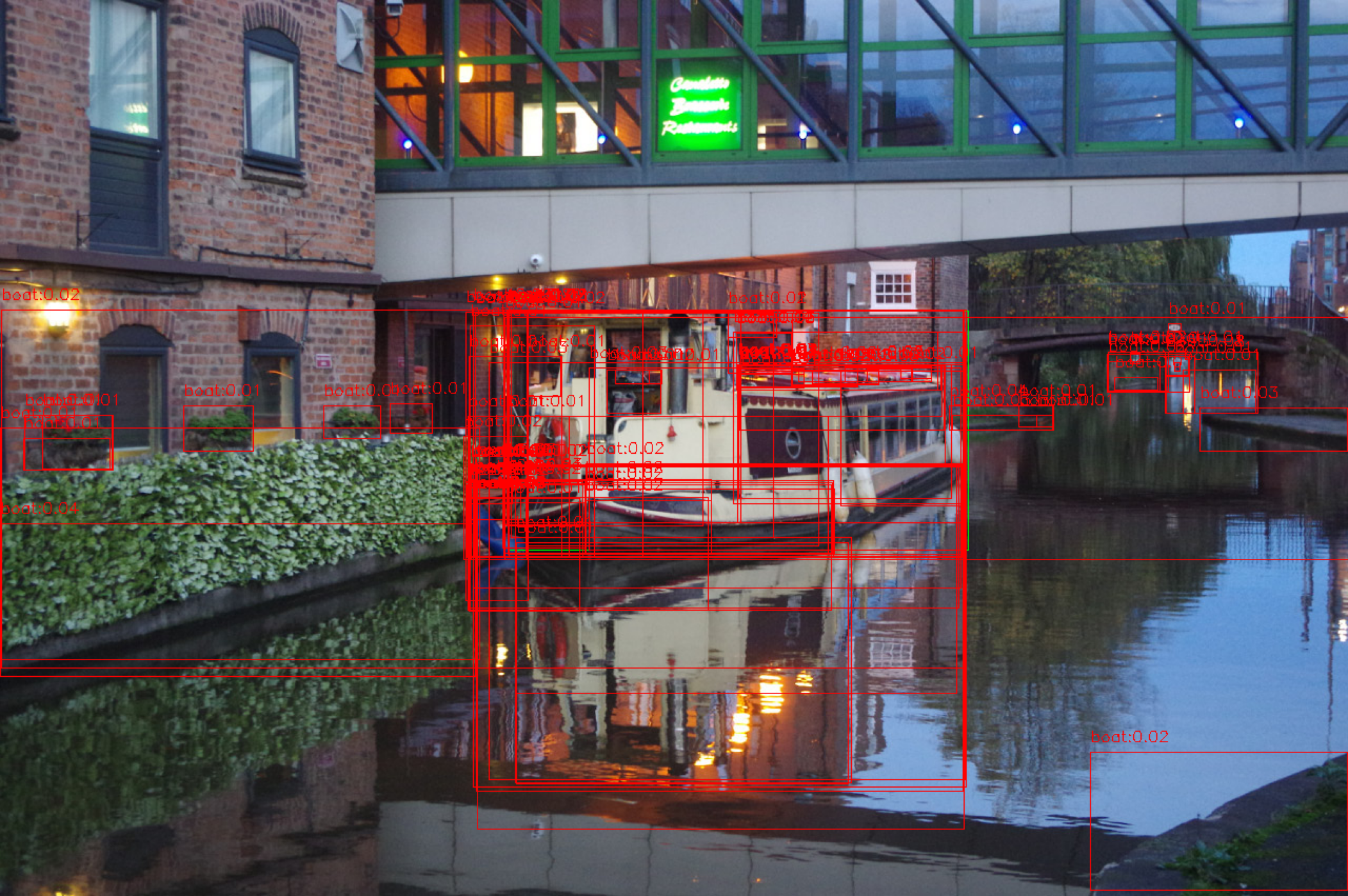}}}&
\bmvaHangBox{\fbox{\includegraphics[width=4.2cm]{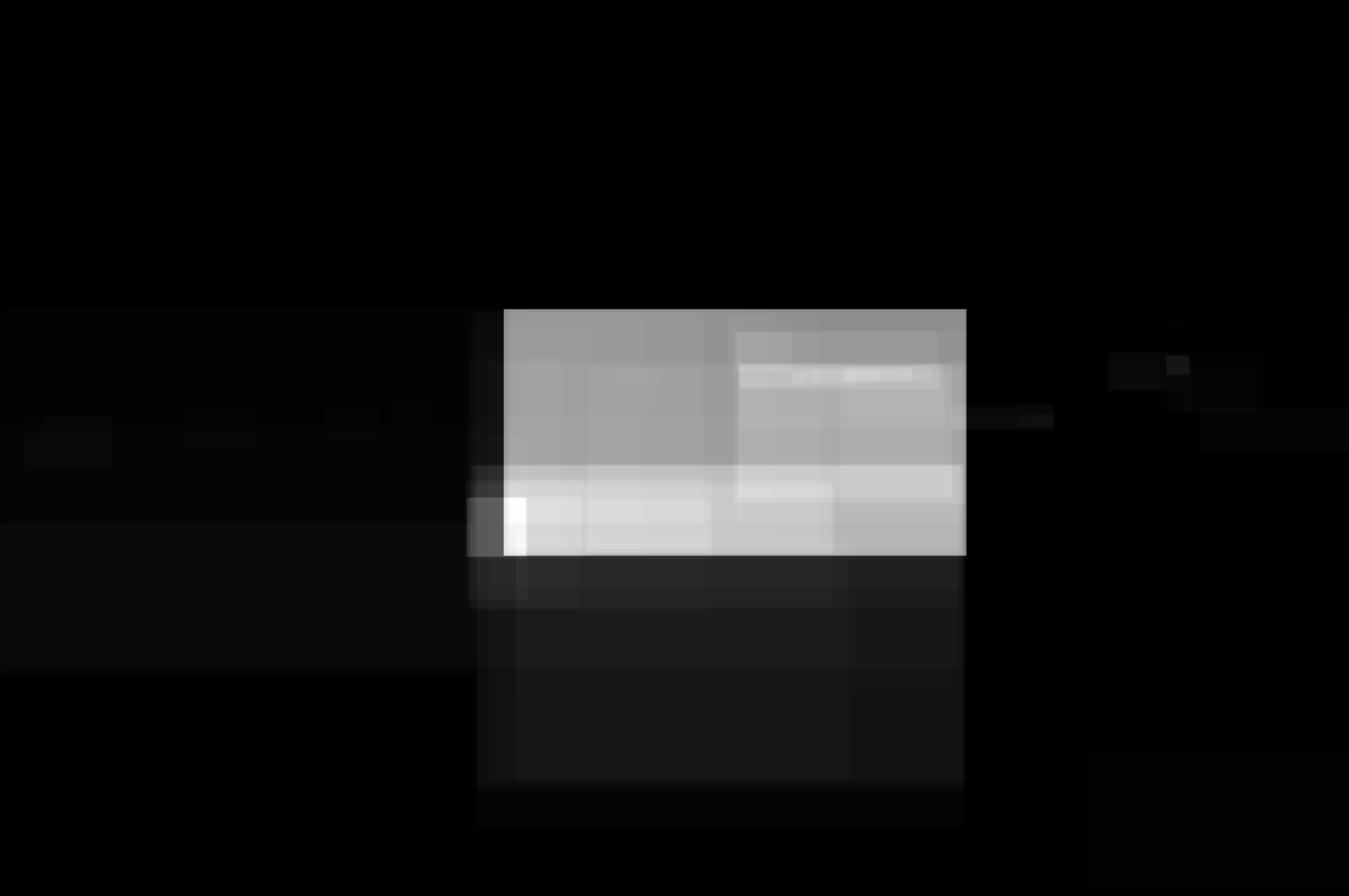}}}\\
(a)&(b)
\end{tabular}
\caption{(a) Proposals on “boat” category; (b) Heatmap of the same category constructed with confidences.
[Source: (a) - Modified from the photo "'L'eau-t Cuisine' on the Shropshire Union Canal at Chester by Stephen McKay" under the same license, CC BY-SA 2.0]}
\label{fig:heatmap}
\end{figure}

We visually assess the proposals of various detectors on our dataset and notice that proposals focus on and group at where the “real” objects are, generally. Inspired by this observation, we aim to utilize this to differentiate between boats and their reflections. We create a so-called “heatmap” using the proposed bounding boxes and their scores. For each pixel in the image, we sum together confidences of the proposals encompassing that pixel. We construct the heatmap class-wise, one heatmap for each class. Visual analysis of these heatmaps shows that real boats are much brighter compared to their reflections, indicating the model is more confident of real objects. This is illustrated in Figure \ref{fig:heatmap}. We formulate a simple approach: If we translate a bounding box upward on the heatmap and see an increase in the mean “heat” inside the box, then this box should be a reflection. We name this approach “Heatmap Based Sliding Filter”. See Algorithm \ref{alg:hbsf} for the exact formulation.

We analyse HBSF on Co-DETR. Results are presented in Table \ref{table:hbsfhbefcomp}.
The mAP is calculated by the mAP tool by Cartucho et al.~\cite{cartucho}. The filter runs on the raw output files of the models, and we were unable to run MMDetection locally. Therefore, we used another tool. The results may slightly differ compared to previous tables. 
Also, we notice false positives have generally lower confidences. Therefore, we apply a small confidence threshold to select candidate proposals for the HBSF filter. We notice HBSF performs better this way. 

\begin{table}[h]
  \rowcolors{2}{gray!25}{white}
  \centering
  \begin{tabular}{cccccc}
    Reflections     & Model           & Filter            & mAP[0.50] & TP@0.5[boat] & FP@0.5[boat]      \\
    Original        & CoDETR          & None              & 0.815     & 483          & 14120             \\
    Inpainted       & CoDETR          & None              & 0.817     & 485          & 12433             \\
    Original        & CoDETR          & Score Thr.[0.3]   & 0.770     & 428          & 344               \\
    Original        & CoDETR          & HBSF              & 0.811     & 475          & 9228              \\
  \end{tabular}      
\caption{\textsc{HBSF Results}}
  \label{table:hbsfhbefcomp}
\end{table}

\begin{figure}[h]
\centering
\begin{tabular}{cc}
\bmvaHangBox{\fbox{\includegraphics[width=6.6cm]{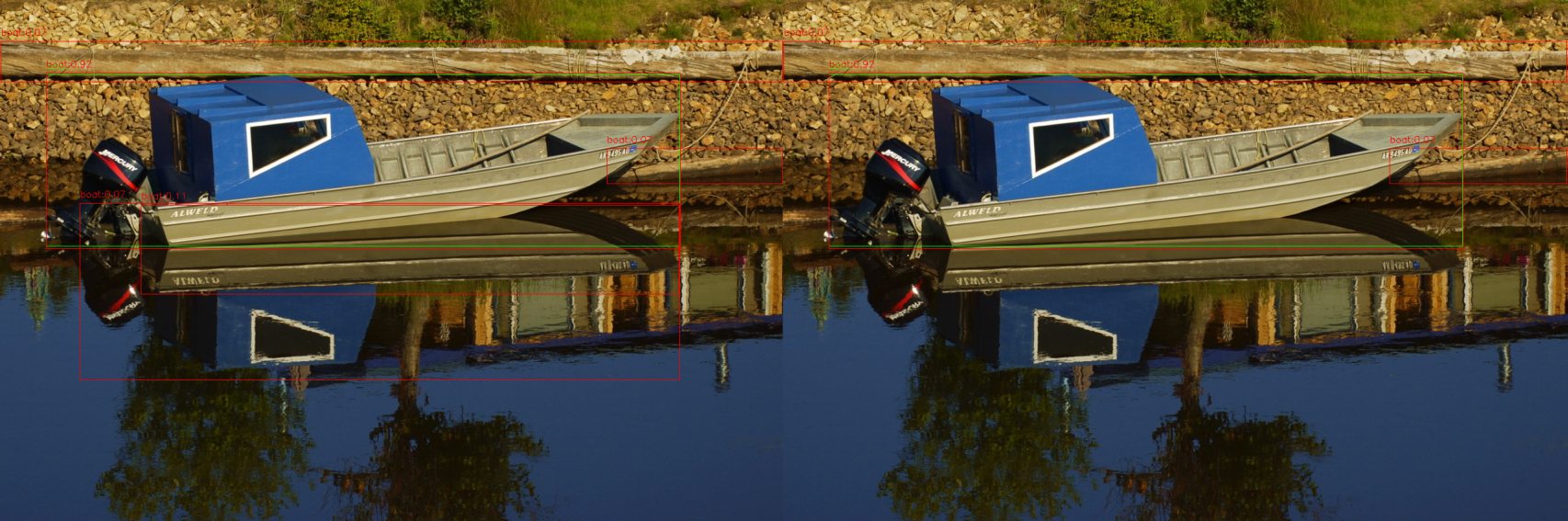}}}&
\bmvaHangBox{\fbox{\includegraphics[width=4cm]{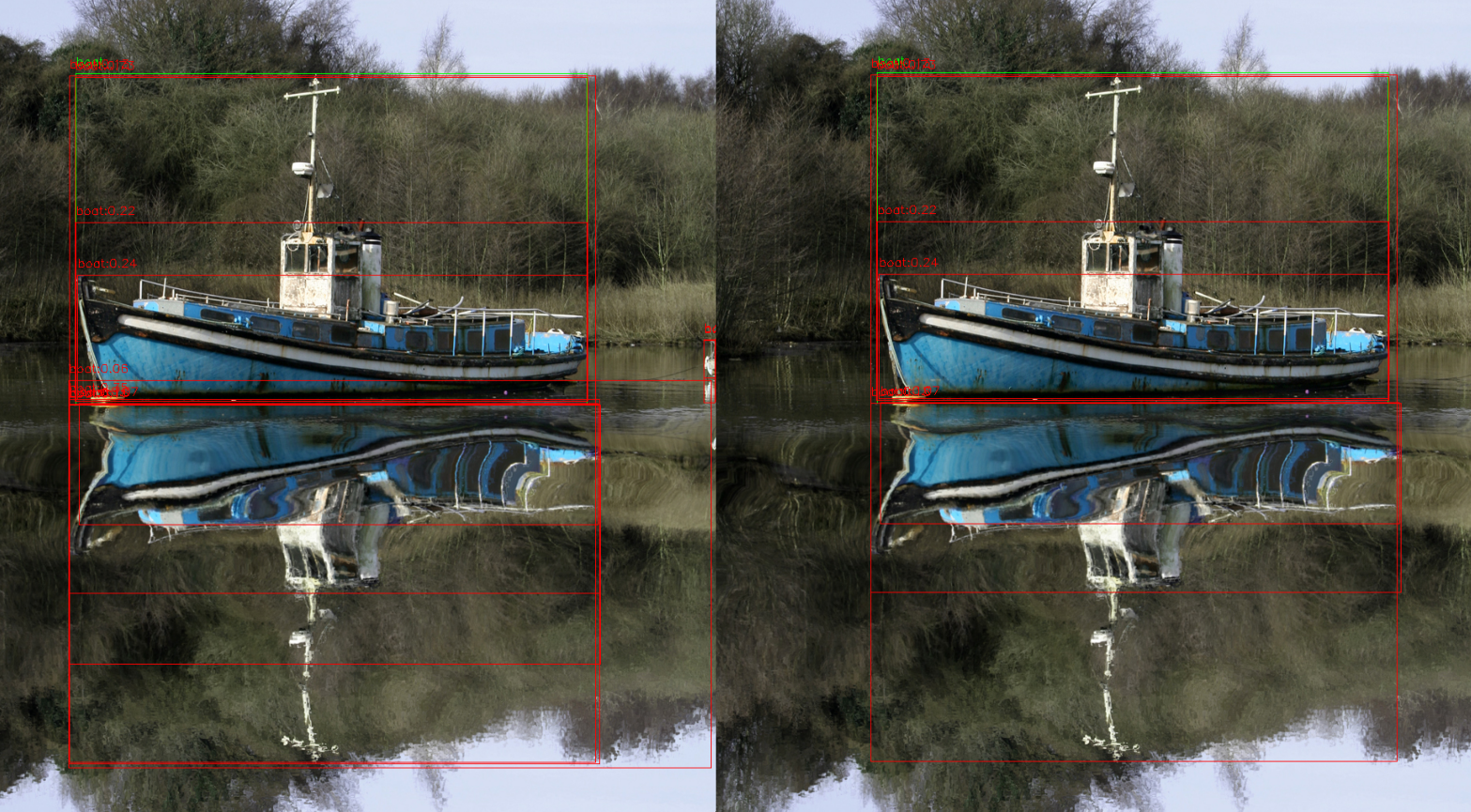}}}\\
(a)&(b)
\end{tabular}
\caption{(a) Proposals before (left) and after (right) HBSF. (b) Example of HBSF failing to remove false positive on reflection.
[Source: (a) - Modified from the photo by Dave Bezaire \& Susi Havens-Bezaire under the same license, CC BY-SA 2.0;
(b) - Modified from the photo by Ben Salter under the same license, CC BY 2.0]}
\label{fig:heatmap_comp}
\end{figure}

We achieve 34.64\% reduction on false positives with the Heatmap Based Sliding Filter (HBSF). The TP count drops by 1.65\%, which is small compared to the reduction in FPs we achieve. We observe a small decrease in mAP, due to missing TPs. These missing TPs come from samples like marinas, where the boats are more cluttered. Figure \ref{fig:heatmap_comp}(a) shows that HBSF successfully filters false positives on reflections. Figure \ref{fig:heatmap_comp}(b) presents a scenario where HBSF fails. 

\begin{algorithm}[h]
\caption{Heatmap-based Sliding Filter}
\begin{algorithmic}[1]
\State Let $r_p^{\uparrow}$ be the region $r_p$ moved up by 1\% of the height
\For{$I \in \mathcal{I}$} \Comment{Iterate over the set of images $\mathcal{I}$}
    \For{$c \in \mathcal{C}$} \Comment{Iterate over the set of classes $\mathcal{C}$}
        \State $H \gets \mathbf{0}$, $\text{dim}(H) \gets \text{dim}(I)$ \Comment{Initialize heatmap $H$ with dimensions of image $I$}
        \For{$p \in \mathcal{P}$} \Comment{Iterate over the set of predictions $\mathcal{P}$}
            \State $H[r_p] \gets H[r_p] + \text{conf}(p)$ \Comment{Add confidence of $p$ to heatmap $H$ at region $r_p$}
        \EndFor
        \For{$p \in \mathcal{P}$} \Comment{Iterate again over the set of predictions $\mathcal{P}$}
            \State $\mu_1 \gets \frac{1}{|r_p|} \sum_{(x,y) \in r_p} H(x,y)$ \Comment{Compute mean $\mu_1$ of heatmap over region $r_p$}
            \State $\mu_2 \gets \frac{1}{|r_p^{\uparrow}|} \sum_{(x,y) \in r_p^{\uparrow}} H(x,y)$ \Comment{Compute mean $\mu_2$ over region $r_p^{\uparrow}$}
            \If{$\mu_2 > \mu_1 \And \text{conf}(p) < 0.3$} \Comment{Compare the mean values $\mu_1$ and $\mu_2$}
                \State $\mathcal{P} \gets \mathcal{P} \setminus \{p\}$ \Comment{Remove $p$ from $\mathcal{P}$ if $\mu_2 > \mu_1$}
            \EndIf
        \EndFor
    \EndFor
\EndFor
\end{algorithmic}
\label{alg:hbsf}
\end{algorithm}
\section{Discussion}
\label{sec:discussion}
The sensitivity analysis presented in Table \ref{table:experiment1} shows that models are sensitive to the presence of surface reflections and identify such reflections as false positives. Visual analysis also suggests that in addition to false positives, reflections cause proposals that encompass both the object and the reflection itself. As can be seen in Table \ref{table:experiment2}, even with domain adaptation, reflections pose a problem. This is in line with findings in MaCVi workshops~\cite{macvi1}~\cite{macvi2}. Specifically, Kiefer et al. observe a reduction in F1 score from approximately 80\% in moderately reflective scenes to below 40\% in highly reflective scenes in the obstacle segmentation challenge.  

The proposed HBSF succeeds in significantly reducing false positives. Specifically, it eliminates 34.64\% of the false positives in our datasets. The qualitative results show that the HBSF eliminates proposals on the reflections. However, the algorithm could be further optimized, as it reduces TPs by 1.65\%. Considering downstream tasks such as path planning for USVs, it is our opinion that this trade off is beneficial. Falsely detecting reflections as obstacles will hinder the autonomous driving capabilities of USVs. It is also important to note that we consider all proposals in our analysis without applying a specific confidence threshold, in order to keep recall as high as possible. As safety is of utmost importance in autonomous driving, achieving fewer FPs at highest recall should be aimed. Typically, object detectors are used with a confidence threshold hyperparameter that eliminates proposals below that threshold. This approach disregards the possibility of low confidence true positive proposals, and therefore reduces recall, thus reducing safety. From Table \ref{table:hbsfhbefcomp} we can see that a typical score threshold of 0.3, although mostly eliminates false positives, reduces the TP by 11.38\%, nearly 7 times of that of HBSF.

Expanding on areas to improve, we note the following: First, we use a value of 1\% to slide the bounding boxes upward. We chose this value as it should sufficiently move the proposals on the reflections upward without causing proposals on the ground truths itself to intersect with other proposals. Secondly, visual assessment of the proposals of Co-DETR prompted us to use a score threshold of 0.3 for the HBSF. In the future, doing a parameter search to optimize these values can possibly improve the filter.

\section{Conclusion}
\label{sec:Conclusion}

In this paper, inspired by previous work~\cite{macvi1}\cite{macvi2}, we analyse the effect of reflections on various detectors for a maritime obstacle detection task. To isolate the reflection variable, we construct a dataset of images with similar perspective to USVs, containing reflections. We also construct a twin of this dataset, where we inpaint the reflections to allow analysis of the impact of reflections in isolation. We make these datasets publicly available. To address the challenge of false positives caused by surface reflections, we introduced the Heatmap Based Sliding Filter (HBSF), a novel approach designed to specifically mitigate the impact of water surface reflections. We demonstrated that our method achieves a 34.64\% reduction in false positives, with only a marginal decrease of 1.65\% in true positive detection, highlighting its effectiveness in distinguishing real obstacles from their reflections.

Our findings underscore the importance of specialized approaches for dealing with reflective surfaces in maritime settings, a domain that has received limited attention in the literature. The results indicate that HBSF is a promising tool for improving the robustness of obstacle detection systems in challenging maritime environments. Moving forward, there is potential to further refine our approach to recover the slight drop in true positive rate, possibly through conducting a parameter search.

\bibliography{references}
\end{document}